\documentclass{article}

\usepackage{arxiv}

\usepackage[utf8]{inputenc} 
\usepackage[T1]{fontenc}    
\usepackage{hyperref}       
\usepackage{url}            
\usepackage{booktabs}       
\usepackage{amsfonts}       
\usepackage{nicefrac}       
\usepackage{microtype}      
\usepackage{graphicx}
\usepackage{natbib}

\usepackage{color}
\usepackage{listings}
\usepackage{caption}

\newcounter{nalg} 
\DeclareCaptionLabelFormat{algocaption}{Algorithm \thenalg} 

\lstnewenvironment{algorithm}[1][] 
{   
    \refstepcounter{nalg} 
    \captionsetup{labelformat=algocaption,labelsep=colon} 
    \lstset{ 
        mathescape=true,
        frame=tB,
        numbers=left, 
        numberstyle=\tiny,
        basicstyle=\scriptsize, 
        keywordstyle=\color{black}\bfseries\em,
        keywords={,input, output, return, datatype, function, in, if, else, foreach, while, begin, end, } 
        numbers=left,
        xleftmargin=.04\textwidth,
        #1 
    }
}{}

\title{A Novel Approach to Lifelong Learning:\\ The Plastic Support Structure}

\author{Georges Kanaan\thanks{The authors acknowledge that they have contributed equally to this paper. The authors would like to thank Professor John Vervaeke for continued wisdom and feedback, and Professor Steve Mann for resourceful support and guidance.} \\
	Department of Computer Science \\
	University of Toronto \\
	Toronto, Canada \\
	\texttt{georges@gkanaan.com} \\
	\And
	Kai Wen Zheng \\
	Department of Computer Science \\
	University of Toronto \\
	Toronto, Canada \\
	\texttt{kevin97illu@gmail.com} \\
	\And
	Lucas Fenaux \\
	Department of Computer Science \\
	University of Toronto \\
	Toronto, ON \\
	\texttt{lucas.fenaux@mail.utoronto.ca} \\
}

\renewcommand{\shorttitle}{\textit{arXiv} The Plastic Support Structure}

\hypersetup{
pdftitle={A Novel Approach to Lifelong Learning: The Plastic Support Structure},
pdfsubject={cs},
pdfauthor={Georges Kanaan, Kai Wen Zheng, Lucas Fenaux},
pdfkeywords={Plasticity, Neural Networks, Multi-task learning},
}

\begin{document}

\maketitle

\begin{abstract}
We propose a novel approach to lifelong learning, introducing a compact encapsulated support structure which endows a network with the capability to expand its capacity as needed to learn new tasks while preventing the loss of learned tasks. This is achieved by splitting neurons with high semantic drift and constructing an adjacent network to encode the new tasks at hand. We call this the Plastic Support Structure (PSS), it is a compact structure to learn new tasks that cannot be efficiently encoded in the existing structure of the network. We validate the PSS on public datasets against existing lifelong learning architectures, showing it performs similarly to them but without prior knowledge of the task and in some cases with fewer parameters and in a more understandable fashion where the PSS is an encapsulated container for specific features related to specific tasks, thus making it an ideal "add-on" solution for endowing a network to learn more tasks.
\end{abstract}

\keywords{Plasticity \and Neural Networks \and Multi-task learning}

\section{Introduction}
The problem of lifelong learning \citep{Thrun95} is that of efficiently representing new tasks given previously known tasks, specifically to leverage known information to improve learning of new information through transfer learning. In the context of our approach, which is that of a deep neural network, we can treat the weights of the network as the encoded information learned by the network which allows us to transfer knowledge by transferring the weights representing said knowledge. 

Many well guided and successful approaches exist, for example recent work has show that using a regularizer we can maintain the knowledge encoded by a set of weights preventing it from drifting too fat as information is learned \citep{kirkpatrick}. one such approach from which our solution is built, is dynamically expandable networks (DEN) \citep{DEN}. DENs achieve lifelong learning by splitting weights that become saturated with information (have a high semantic drift) and restoring the previous weights. The new weights are copied into new neurons, then trained to better represent the new task. If any additional new neurons are required, they will be added and trained as well on the current task. This results in the network successfully containing a representation for the current task alongside the previously successfully learned information.

PSS is inspired by the splitting mechanics of DEN but differs greatly in how the copied neuron is added in the network, as well as how the network is trained. Most importantly, PSS does not need to have a task specified with the current input, that is, instead of executing binary classification of an input with respect to a specific task, PSS allows a network to confidently classify an input as one of the learned tasks. In addition, PSS opts out of using a timestamp mechanism like DEN's, which associates each neuron with the tasks it has been trained on. Instead PSS offers a more streamlined and contained approach granting the network greater flexibility in its learning and structure while reducing overhead. 

The challenges our approach faces are essentially that of lifelong learning in general, first we would like to avoid catastrophic forgetting, that is forgetting previously seen tasks in an unrecoverable way, drastically affecting future performance regarding those tasks. Second, we would like to avoid exponential growth of the amount of neurons leading the network to simply over fit all seen data. Finally, we would like the network to be efficient.

\section{Dynamically Expandable Networks}
To understand the approach we have taken, one must first look at dynamically expandable networks (DEN). Unlike Progressive Networks \citep{Rusu} which entirely prevents modifications to the existing structure, DENs dynamically expand their capacity for information by adding a set of blank neurons to each layer, exclusively training them, then pruning the neurons who are said to have no impacts on the network based on the value of their weights \citep{DEN}. They tackle the issue of catastrophic forgetting by training a neuron on the current task then calculating its \emph{semantic drift} between the weights current value and its value prior to the current task, if this value is above a certain threshold defined empirically, then the node is reverted to its previous value and copied, the copy is then exclusively trained. The issue in this approach is that, collectively, the new weights drastically affect the outputs of the nodes in the next layer which due to \emph{selective retraining} have not had a chance to adapt to the addition of these nodes, we introduce this problem as the influencing weights problem. DENs solve this with \emph{timed inference}, i.e. marking each neuron with the tasks it affects, then artificially restricting the network to only neurons related to the task being passed in as input \citep{DEN}. Hence DEN relies on knowing the task associated with the input to produce a prediction as to whether the input belongs or not to the task.

\section{Plastic Support Structure}

\begin{figure}
\centering
\caption{An example neural network with the PSS structure in red.}
\includegraphics[width=70mm]{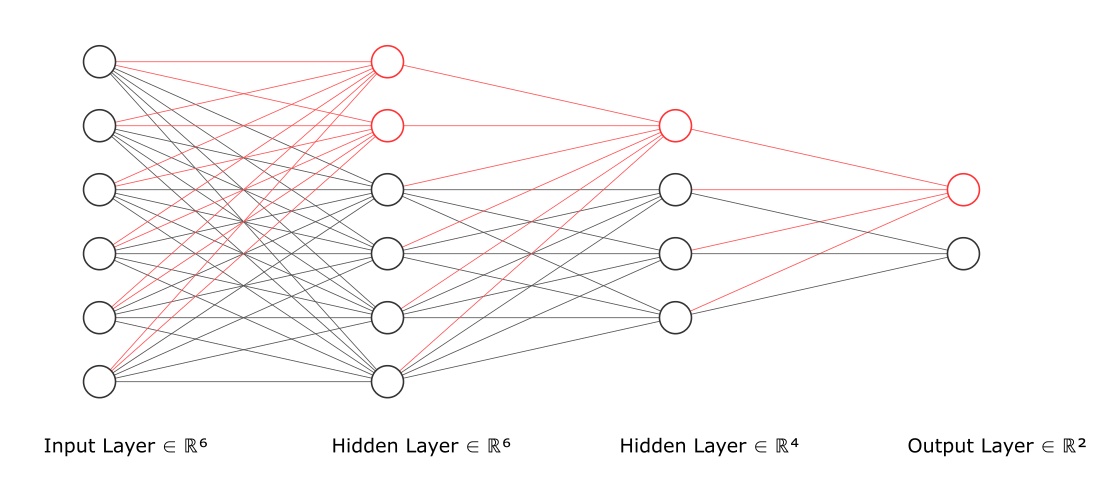}
\label{fig:nn1}
\end{figure}

The plastic support structure seeks to solve the influencing weights problem in a more natural and flexible solution which allows the network to maximize the information it encodes. In addition, unlike DEN, the PSS would like to do so without the knowledge of which task is being classified once training is complete. That is, we would like the network to confidently identify which task a given input belongs to, rather than confirming whether an input belongs to a given task. Our network is implemented in python using pytorch \citep{Pytorch} as the main machine learning library, code can be found on any of the author's GitHub profile. Figure \ref{fig:nn1} shows in red the structure of a PSS, the red links are those that affect the PSS only.

\subsection{Expansion and Support Structures}
\begin{algorithm}[caption={SPLITTING}, label={alg1}]
INPUT: $W_t$ and threshold $\sigma$
for all hidden units i do:
  drift $\gets$ $||W^t_i - W^{t-1}_i||^2$
  if drift > $\sigma$:
    create new node $n$ for the layer.
    copy incoming weight from $i$ to $n$
    set outgoing weights of $n$ connected to old nodes to 0

for hidden layers which did not receive a new node $l$:
  add new node to $l$
\end{algorithm}

The plastic support structure retains the same criteria for splitting as a DEN. That is, for each neuron, given the weights at task $t$ and task $t-1$, we calculate the semantic drift between them which is the norm of the difference between both weights, if the drift is greater than an empirically defined threshold (the drift threshold) we revert the neuron to its value at task $t-1$ and duplicate it. However in PSS, a split neuron must have a path exclusively across new neurons to the output neuron, hence if one neuron is split, we must at least expand each subsequent layer by one neuron. This expansion has the benefit of increasing the network's capacity for information. Note that the drift threshold is increase by another per-layer parameter, the drift delta, every epoch in order to better manage the evolution of the model over time.
 
In order to solve the influencing weights problem, we isolate these new neurons' outputs from the old neurons in the next layer of the network, that is the weight of the output of each new neuron to an old neuron is zero, unless the old neuron is an output neuron. The network can still access the support structure at every layer since the input weights of the neurons are unmodified, simply put, the output of new neurons won't affect old neurons, instead it is propagated forward within the PSS until it reaches the output neuron. 
Our theory is that this method will provide the network with the capacity to encode tasks highly efficiently. 

\subsection{Selective Retraining}
\begin{algorithm}[caption={SELECTIVE RETRAINING}, label={alg2}]
INPUT: Dataset $D_t$, threshold $\sigma$:
for all weights in network w:
  if $|w|$ < $\sigma$:
    freeze $w$.
train network.
\end{algorithm}

Much like DEN we employ a version of selective retraining in order to train parts of our network exclusively \citep{DEN}. Where a DEN employs selective retraining by neuron, we employ it by weight. That is, instead of freezing entire neurons, our implementation of selective retraining allows for subsets of a neuron's weights to be frozen. We define freezing a weight by setting its gradient to zero during back propagation so that the value of the weight remains unchanged.
We employ selective retraining by weight because we would like to train the inputs of neurons in the PSS from every neuron in the previous layer, without said neurons affecting other neurons outside the PSS. Selective retraining deals with exclusively allowing for neurons in the PSS to be trained without affecting the values of nodes outside of it, ensuring the integrity of the network's structure and that no old task's output is modified by addition of the PSS.

In addition, we make use of a truncated approach to back propagation within selective retraining which accounts for the fact that only weights related to task being trained will be updated. We do not seek to truncate back propagation through time as in an RNN \citep{truncatedbackprop}. Instead we seek to endow selective retraining with an efficient method of updating the weights relative only to a certain task. To do this we back propagate solely through the output nodes of interest, in that sense we truncate back propagation from executing on all output nodes to a selected subset. This allows for more efficient training cycles where we can guide isolate nodes that have no effect on the current output from being updated and preserve computational power and time.

\section{Experiments}
In our experiments, we mainly compared our model to the DEN \citep{DEN} as it was the model we decided to improve on.
To do so, we used two datasets:

- MNIST: The MNIST dataset \citep{mnist} which consists of a training set of 60,000 images and a test set of 10,000 images. We used all 60,000 images for training and 2,000 images for testing. 

- MNIST variation: The MNIST dataset \citep{mnist} with an added Gaussian noise of mean 0 and standard deviation 1. We also used the 60,000 images for training and the 2,000 images for testing.

For the DEN, we used the configuration and results from \citep{DEN} which was a two-layer feedforward network with layer sizes 312 and 128.

We decided to use a feedforward network with the following layout: two-layers with 312 and 128 neurons each with LeakyRelu activation functions \citep{relu}. These were implemented using the Pytorch \citep{Pytorch} library. Our code will be published upon acceptance of the paper. The layer sizes were decided in order to enable fair comparison between our model and the DEN results.

We achieved a final accuracy of 95.05\% average accuracy over all tasks on the MNIST dataset, and a final accuracy of 80.50\% on the MNIST variation dataset. 

The added noise in MNIST variation makes the dataset more challenging to learn relative to the standard MNIST dataset. Throughout our testing, we noticed that the drift threshold and drift delta parameters have a noticeable influence on the evolution of the model's structure and hence it's results. An increasing drift threshold through a drift delta that was often as great as the original drift threshold was found to work best to limit exponential growth in the inner layers and keep the model spatially efficient. For example, on MNIST variation, we ended up using 0.3 and 0.25 for the initial drift thresholds for the inner layers and 0.0015 and 0.25 for the drift deltas. However, it is worth noting, as seen in the values above, that the behavior from layer to layer can be drastically different, since the first layer needed a much smaller drift delta to prevent exponential growth compared to the second layer.

This could be due to the general purpose that each of the layers usually serve in a feedforward network when working with images. The first inner layer identifies features in the images while the second inner layer refines those features and maps each feature to it's corresponding task weighted with it's importance to that task. Once the feature map is built, it will not change much as the tasks progress. It is also worth noting that since we are truncating our output before executing the back-propagation, the second inner layer connected to each task's output node is only trained once that task is reached, whereas the first inner layer is trained throughout all the tasks.

This feature of our model can allow for additional tasks to be added during the training as well, as all it would require would be to add an extra output node and fully connect it to the last inner layer and those extra connections wouldn't have "missed" anything since they wouldn't have been trained until their task was reached anyways.

Therefore when using our model, it is usually worth using a fairly small drift delta for all layers except the last one, where we would want the drift delta to be far greater.

\begin{figure}
\centering
\caption{Per-task performance of the DEN and the PSS models on MNIST-Variation}
\includegraphics[width=70mm]{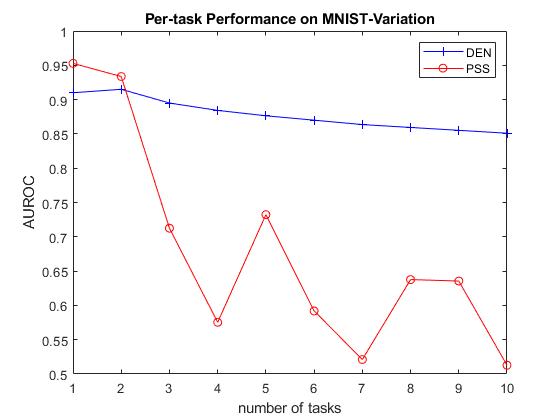}
\label{fig:plot1}
\end{figure}

As shown in figure \ref{fig:plot1}, it is challenging for the model to learn each subsequent task. We can see that our network is not competitive enough yet to achieve performances on par with the DEN or state of the art networks. However, we believe that with more improvements and added support for types of networks other than neural nets, our support system could bring a significant improvement to lifelong learning networks.

\section{Conclusion}
The plastic support structure is a simple, well-encapsulated solution to endow a network with plasticity. It has few parameters, assumes no prior knowledge of the task being classified, and performs well as we have seen on MNIST and a variation of the MNIST dataset. The PSS performs similarly to existing solutions with the added benefits previously mentioned, and in some cases can simplify an existing architecture and model. PSS also does not store any knowledge of tasks on a per-node basis, allow the model to efficiently use it's entire capacity to encode the necessary information, and freeing it from requiring prior knowledge. The rate of growth of neurons or "plasticity" of the network can be tailored using the drift threshold making the PSS suitable for environments where resources may be limited such as mobile devices, it allows the architect of the model to decide how much space to be granted for additional tasks for any reason.

Future work might consist of making the PSS more efficient by more closely studying the way links are formed within it and with the rest of the model. For example, the current structure precludes the PSS from augmenting existing knowledge in the hidden layers of the rest of the model, since it's forward connections are solely within itself. Another area of interest could be that of well-defining an optimal drift threshold, or maybe multiple thresholds, juggling between information learnt and resources used. This would have the added benefit of yet again reducing complexity of the parameters added by the PSS and render it more accessible and easy to use in it's a role as an "add-on plasticity module". Finally, we'd like to see work leading to an increase in per-task accuracy more closely resembling what we see in the state-of-the-art.

The PSS is great tool to easily endow an arbitrary architecture with plasticity, it's relatively simple, accurate and has a low-impact on existing network given the way it's connections are formed internally. We hope that it will spur more interest in the topic of plasticity in neural networks.  

\bibliographystyle{unsrtnat}
\bibliography{main}

\end{document}